\documentclass[10pt,twocolumn,letterpaper]{article}

\usepackage{iccv}
\usepackage{times}
\usepackage{epsfig}
\usepackage{graphicx}
\usepackage{amsmath}
\usepackage{amssymb}
\usepackage{float}
\usepackage{caption}
\usepackage{subcaption}
\usepackage{xcolor}
\usepackage{multirow}

\usepackage[pagebackref=true,breaklinks=true,letterpaper=true,colorlinks,bookmarks=false]{hyperref}

\iccvfinalcopy 

\def\iccvPaperID{9} 

\ificcvfinal\fi
\begin{document}

\title{ShelfNet for Fast Semantic Segmentation}
\author{Juntang Zhuang$^1$, \quad Junlin Yang$^1$,\quad  Lin Gu$^2$ \quad Nicha C. Dvornek $^{1}$\\
$^{1}$ Yale University, USA  \\ $^{2}$ National Institute of Infomatics, Japan \\
{\tt\small \{j.zhuang; junlin.yang; nicha.dvornek;\}@yale.edu, ling@nii.ac.jp}
}

\maketitle

\begin{abstract}
In this paper, we present ShelfNet, a novel architecture for accurate fast semantic segmentation. Different from the single encoder-decoder structure, ShelfNet has multiple encoder-decoder branch pairs with skip connections at each spatial level, which looks like a shelf with multiple columns. The shelf-shaped structure can be viewed as an ensemble of multiple deep and shallow paths, thus improving accuracy. We significantly reduce computation burden by reducing channel number, at the same time achieving high accuracy with this unique structure. In addition, we propose a shared-weight strategy in the residual block which reduces parameter number without sacrificing performance. Compared with popular non real-time methods such as PSPNet, our ShelfNet achieves 4$\times$ faster inference speed with similar accuracy on PASCAL VOC dataset. Compared with real-time segmentation models such as BiSeNet, our model achieves higher accuracy at comparable speed on the Cityscapes Dataset, enabling the application in speed-demanding tasks such as street-scene understanding for autonomous driving. Furthermore, our ShelfNet achieves 79.0\% mIoU on Cityscapes Dataset with ResNet34 backbone, outperforming PSPNet and BiSeNet with large backbones such as ResNet101. Through extensive experiments, we validated the superior performance of ShelfNet. We provide link to the implementation \url{https://github.com/juntang-zhuang/ShelfNet-lw-cityscapes}.
\end{abstract}

\section{Introduction}
\label{introduction}
Semantic segmentation is the key to image understanding \cite{everingham2015pascal,mottaghi2014role}, and is related to other tasks such as scene parsing, object detection and instance segmentation \cite{lin2014microsoft, zhou2017scene}. 
The task of semantic segmentation is to assign each pixel a unique class label, and can be viewed as a dense classification problem. Recently many convolutional neural networks (CNN) have achieved remarkable results on semantic segmentation tasks \cite{long2015fully,chen2016deeplab,zhao2017pyramid}. However, the success of most deep learning models for semantic segmentation comes at a price of heavy computation burden. The training of CNNs on a large dataset such as PASCAL VOC \cite{everingham2015pascal} and Cityscapes \cite{cordts2016cityscapes} typically takes several days on a single GPU, and the running time during test phase is usually hundreds of milliseconds (ms) or more, which hinders their application in time-efficiency demanding tasks.


\begin{figure*}[t!]
    \centering
    \begin{minipage}{.7\textwidth}
        \centering
        \includegraphics[width=0.8\linewidth]{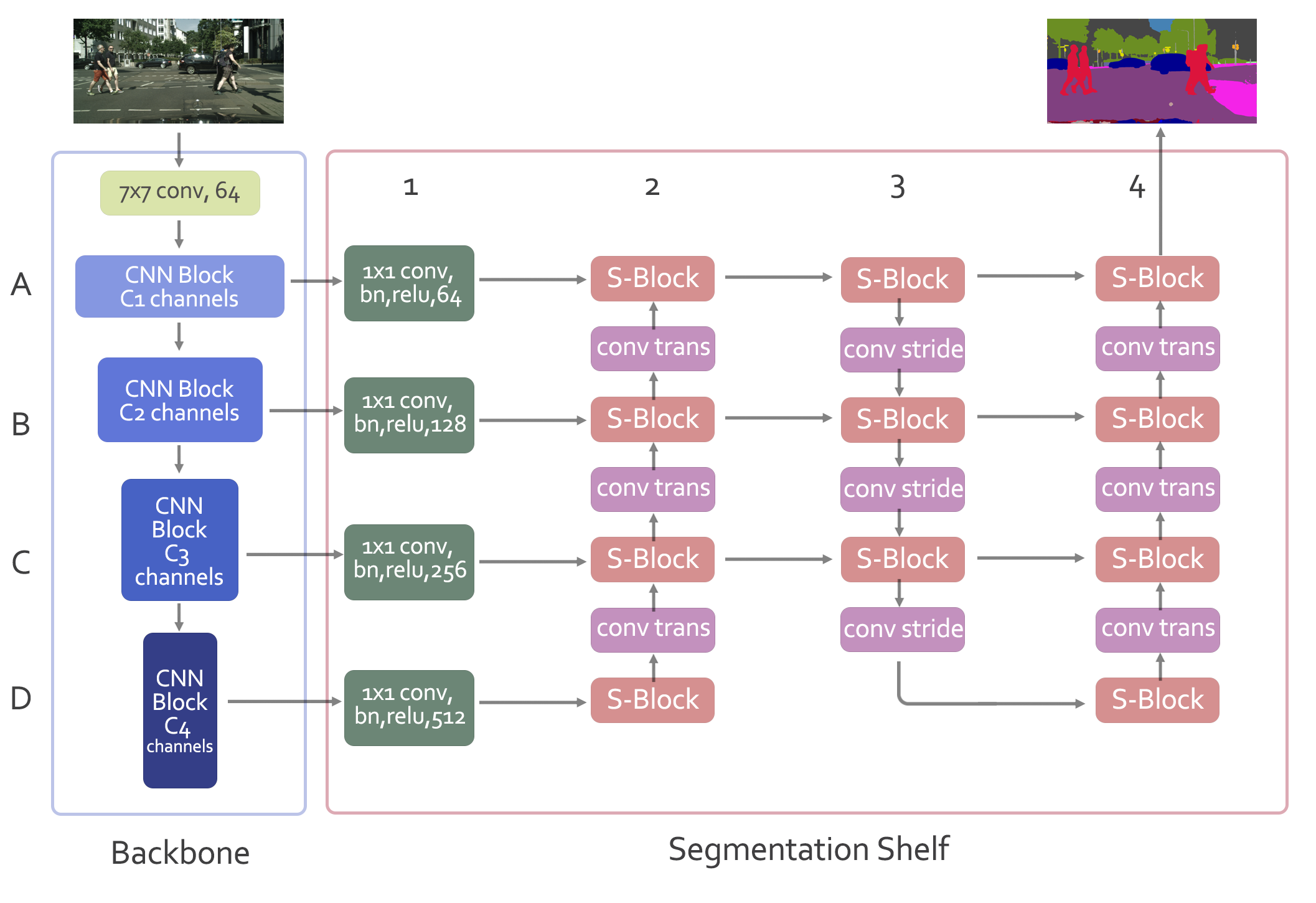}
        \caption*{\small{(a) Structure of ShelfNet.}}
    \end{minipage}%
    \begin{minipage}{0.25\textwidth}
        \centering
        \includegraphics[width=\linewidth,height=6cm]{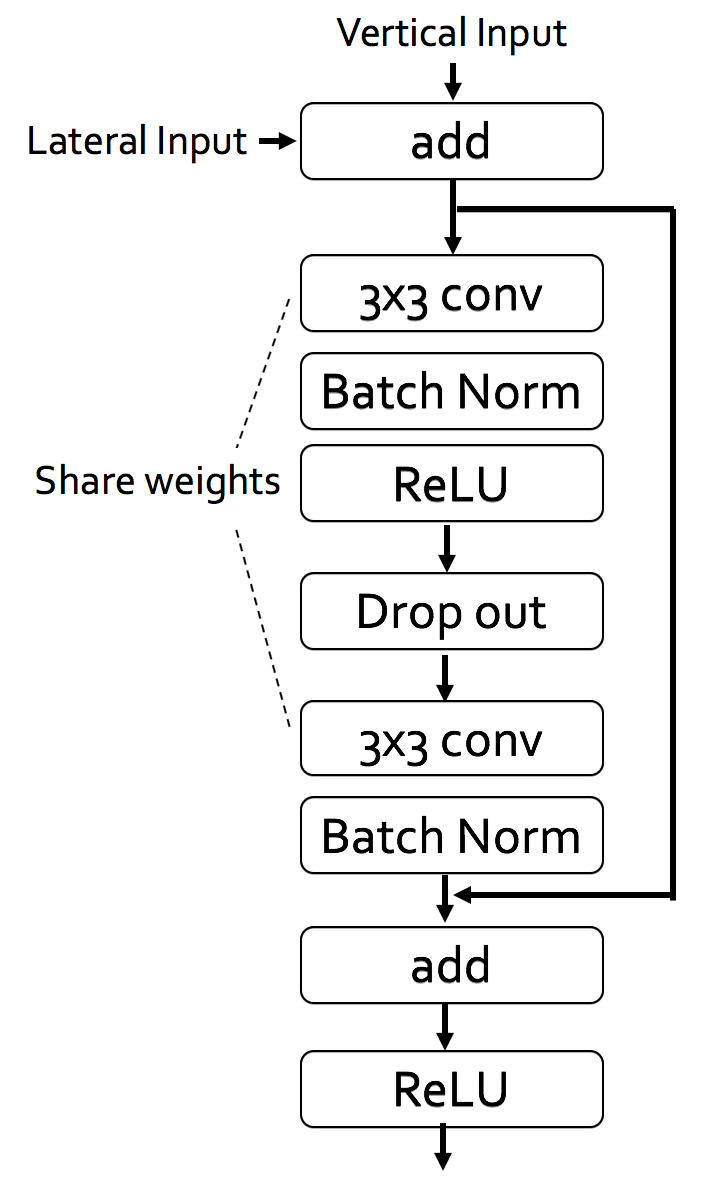}
        \caption*{\small{(b) S-Block (Shared-weight residual block).}}
    \end{minipage}
    \caption{\small{Structure and modules of the ShelfNet. (a) Architecture of ShelfNet. Rows A-D represent different spatial levels (e.g. for ResNet backbone, the spatial sizes of A-D are 1/4, 1/8, 1/16 and 1/32 of input image respectively). Columns 1-4 represent different branches: 3 is ``encoder" (down-sample) branch while 2 and 4 are ``decoder" (up-sample) branches; column 1 reduces the number of channels by $1\times1$ convolution followed by batch normalization and ReLU activation, and the numbers in column 1 represent the channel number of feature maps. Features encoded by stages of backbone (column 0, \textit{i.e.} ResNet) are fed into segmentation shelf. (b) Shared-weight residual block (S-block) where two convolution layers share weights. The shape of feature maps are modified by transposed convolution layer (conv trans) and convolution layer with stride 2 (conv stride). Input and output of a single S-block has the same shape; in the case of two inputs, they are first summed up and fed into S-block.} }
    \label{fig:shelfnet}
\end{figure*}


Real-time semantic segmentation has important applications, e.g., street scene understanding and autonomous driving. Prior research on accelerating semantic segmentation includes removing redundancy of deep neural networks through pruning \cite{han2015deep,han2015learning,hassibi1993second} and distillation \cite{papernot2016distillation,hinton2015distilling,romero2014fitnets}. However, the running speed of these methods is typically insufficient for fast semantic segmentation. Another way to get faster running speed is to use a smaller model keeping the same structure, but this strategy  would inevitably yield lower accuracy \cite{zhao2017icnet}. Therefore, we aim to propose a new architecture for fast segmentation while maintaining satisfying accuracy.

  
Most state-of-the-art semantic segmentation models belong to the family of single ``encoder-decoder" structure, where the image is progressively down-sampled then up-sampled. Here, we propose ShelfNet, which employs a different structure like a multi-column shelf. As illustrated in Fig.~\ref{fig:shelfnet}(a), multi-scale features encoded by different stages of a CNN backbone (\textit{e.g.} ResNet) are fed into the ``\textbf{segmentation shelf}''. The segmentation shelf comprises of  multiple encoder-decoder pairs with skip connections at each spatial resolution level. The unique structure increases the number of paths to improve information flow in the network, thus increasing the segmentation accuracy. We demonstrate the high accuracy and fast running speed of our ShelfNet on PASCAL VOC, PASCAL Context and Cityscapes datasets. 



Our main contributions are listed as follows:

1. We propose a novel architecture, ShelfNet (Fig.~\ref{fig:shelfnet}), for accurate and fast semantic segmentation. Our ShelfNet has a shelf structure of multiple encoder-decoder pairs to improve information flow in the network.

2. In the segmentation shelf, we propose to share the weights of two convolutional layers in the same residual block. This strategy greatly reduces the number of parameters without sacrificing accuracy.

3. We validate the superior performance of ShelfNet on various benchmark datasets. On the Cityscapes dataset, our real-time ShelfNet with ResNet18 backbone achieved 75.8\% mIoU at a comparable speed to BiSeNet; our non real-time ShelfNet with ResNet34 achieves 79.0\% mIoU, outperforming PSPNet and BiSeNet with larger backbones such as ResNet101.




\section{Related Work}
\subsection{Semantic Segmentation}
Semantic segmentation has been a hot topic for many years. Before the recent rise of deep learning, early approaches mainly relied on handcrafted features such as HOG \cite{dalal2005histograms} and SIFT \cite{lowe2004distinctive}. Since the resurgence of deep learning, especially fully convolutional neural networks (FCN) \cite{long2015fully}, deep learning models have been widely used for semantic segmentation.

Like FCN, many models such as U-Net \cite{ronneberger2015u}, RefineNet \cite{lin2017refinenet} and SegNet \cite{badrinarayanan2015segnet} also have an encoder-decoder structure, and use a ``convolution-downsample" strategy for the encoder. The downsample layer reduces spatial resolution and increases channel number. 
  
 Chen et al. proposed DeepLab \cite{chen2016deeplab} based on dilated convolution. Instead of down-sampling, a dilated CNN gradually increases dilation rate to increase the size of receptive field, but does not shrink the size of the output tensor. Therefore, compared with the ``conv-downsample" strategy, the dilated CNN has a better spatial resolution at a higher computation cost. For example, a standard ResNet shrinks image size to $1/32$ of input size, while a dilated ResNet shrinks image size to $1/4$ of input size. State-of-the-art networks, such as DeepLab\cite{chen2016deeplab}, PSPNet \cite{zhao2017pyramid} and EncNet \cite{zhang2018context}, are based on dilated CNN, and thus are not suitable for time-efficiency demanding tasks.
 
\subsection{Real-time Semantic Segmentation}

There have been several approaches for real-time semantic segmentation by modifying a large network to a light-weight version. For example, ICNet \cite{zhao2017icnet} is a modification of PSPNet and deals with multiple image scales, but the robustness to low-resolution is not extensively validated. Light-Weight RefineNet is a modification of RefineNet \cite{nekrasov2018light}, where the kernel sizes of some convolutional layers are reduced from $3\times3$ to $1\times1$. However, most of the real-time models, such as GUN \cite{mazzini2018guided}, EffConv \cite{romera2017efficient} and ENet \cite{paszke2016enet}, achieve high running speed at the cost of low accuracy. The recently proposed BiSeNet
\cite{yu2018bisenet} uses a shallow branch to capture spatial information, and a deep branch to capture context information. However, it's difficult for BiSeNet to deal with high-level features combined from different branches. All models mentioned in this paragraph can be viewed as modifications of the encoder-decoder structure. In this project, we propose a network with multiple encoder-decoder pairs (e.g. columns 3 and 4 in Fig. \ref{fig:shelfnet}) and skip connections at different spatial levels (e.g. rows A-D in Fig. \ref{fig:shelfnet}), and demonstrate its superior performance over previous methods both in inference speed and segmentation accuracy.

\section{Methods}
\label{methods}
\subsection{Structure of ShelfNet}
We propose ShelfNet, a multi-column convolutional neural network for semantic segmentation as shown in Fig. \ref{fig:shelfnet}. Different from the standard single encoder-decoder structure, our shelf-structure network introduces more paths to improve the information flow. 
  
As shown in Fig~\ref{fig:shelfnet}, our ShelfNet relies on a backbone network. Different CNN architectures can be used as the backbone, such as ResNet \cite{he2016deep}, Xception \cite{chollet2017xception} and DenseNet \cite{huang2017densely}. The backbone outputs feature maps of different spatial scales, named as rows A to D in Fig.~\ref{fig:shelfnet}. Take ResNet backbone for example, the spatial sizes of feature maps are 1/4, 1/8, 1/16, 1/32 of the input size at levels A-D, respectively. Feature maps encoded by different stages of the backbone are fed into the segmentation shelf. 
  
The segmentation shelf has a shelf-shaped multi-column structure where columns are named with numbers 1 to 4. We name column 3 as an encoder branch (down-sample branch), and name column 2 and 4 as decoder branches (up-sample branch). Between adjacent columns (e.g. column 3 and 4), there are skip connections at different spatial scale stages (A-D). 
  
Stage-wise features encoded by column 1 are then passed to succeeding Shared-weight blocks (S-block) in column 2. The S-Block serves as a residual-block but with fewer parameters. Here, the S-block combines the features passed from the vertical direction and lateral direction. As shown in Fig.~\ref{fig:shelfnet}(b), the two inputs are first summed up before feeding into the succeeding part. For the exception at A3 and D4 where there is only one input, the summing up step is skipped. As a residual-block, the input and output of a S-block has the same shape.
  
For the encoder branch at column 3, the feature maps are passed into a convolution layer with a stride of 2 (conv stride in Fig.~\ref{fig:shelfnet}), to halve the spatial size and double the channel number (e.g. B3 to C3). Similarly, in the decoder branches at column 2 and 4, feature maps are passed into a transposed convolution layer (conv trans in Fig.~\ref{fig:shelfnet}) with a stride of 2, to double the spatial size and halve the channel number (e.g. C2 to B2). Finally, output from block $A4$ goes through a $1\times1$ convolutional layer and a softmax operation to generate the final segmentation.

\subsection{Channel Reduction for Faster Inference Speed}
\label{sec:channel_reduction}
The backbone feature maps typically have a large channel number (e.g. 2048 at level D for ResNet50). For faster inference speed, we reduce the number of feature map channels with $1\times1$ convolution followed by a batch-norm layer and ReLU activation (e.g. for ResNet50, channel number of feature maps are reduced from 256, 512, 1024, 2048 to 64, 128, 256, 512 for levels A-D respectively). As shown in column 1 in Fig. \ref{fig:shelfnet}, channel number is reduced by a $1\times1$ convolution, batch-normalization and ReLU activation. 

We theoretically show that reducing channel number significantly improves inference speed. The computation burden of a convolution with stride 1 is $H \times W \times K^2 \times C_{in} \times C_{out}$, where $H, W$ are spatial sizes, $K$ is the kernel size, and $C_{in}$ are $C_{out}$ are channel number of input and output respectively. Simply reducing $C_{in}$ and $C_{out}$ by a factor of 4 \textit{quadratically} reduces the computation burden to $\frac{1}{16}$. Even with two extra columns in the shelf structure, the computation burden is $2\times \frac{1}{16} = \frac{1}{8}$, thus is faster.







\begin{figure}[!htb]
    \centering
    \includegraphics[width=1\linewidth]{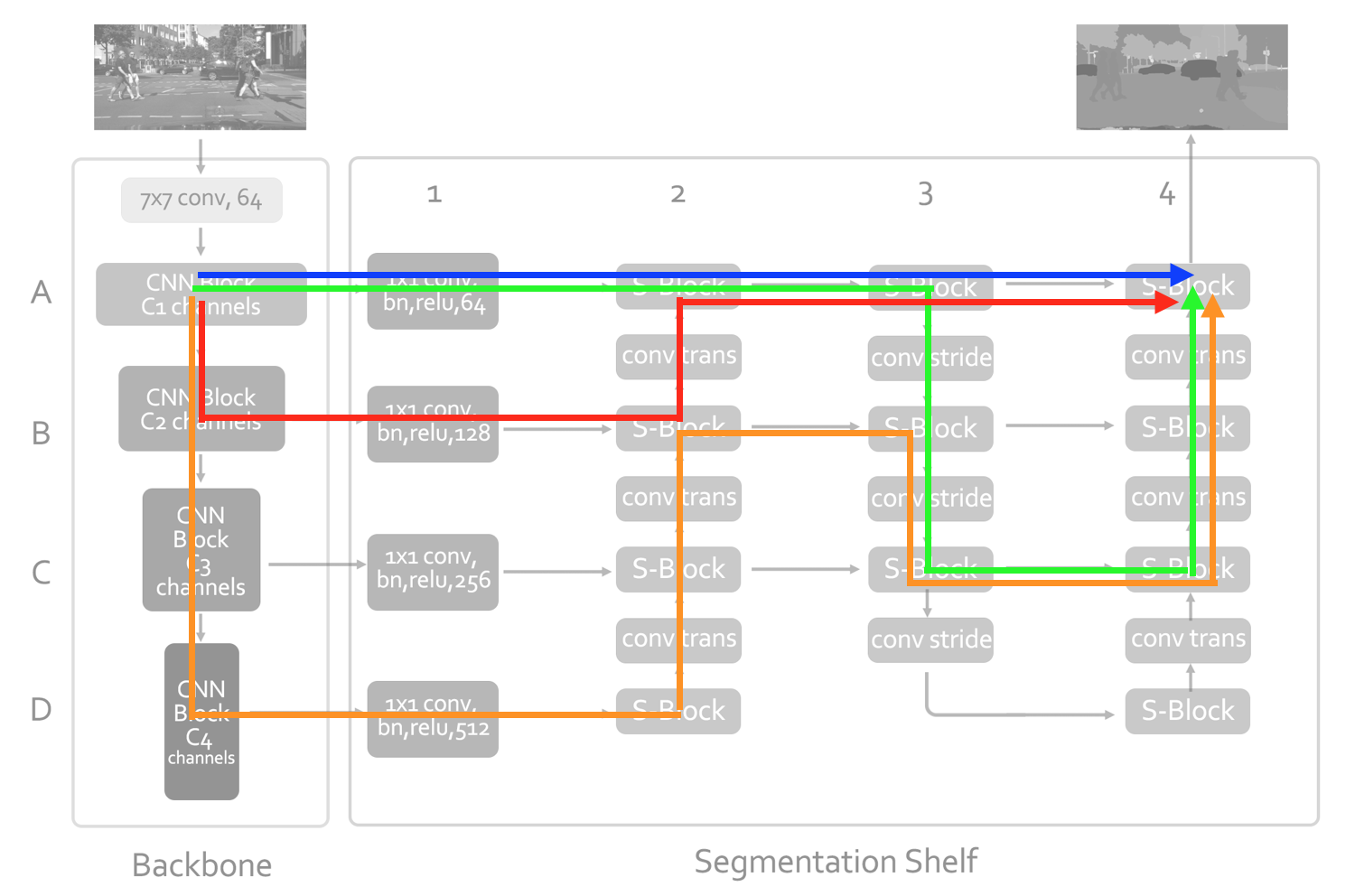}
        \caption{\small{ShelfNet (gray background, the structure is the same as Fig. \ref{fig:shelfnet}) can be viewed as an ensemble of FCNs. A few examples of information flow paths are marked with different colors. Each path is equivalent to an FCN (except that there are pooling layers in ResNet backbone). The equivalence to an ensemble of FCNs enables ShelfNet to perform accurate segmentation.}
        }
    \label{ensemble}
\end{figure}



\subsection{ShelfNet as an Ensemble of FCNs}
\label{sec:fcn_ensemble}

ShelfNet can also be viewed as an ensemble of FCNs. Andreas \textit{et al.} \cite{veit2016residual} argued that ResNet behaves like an ensemble of shallow networks, because the residual connections provide multiple paths for efficient information flow.  Similarly, ShelfNet provides multiple paths of information flow. For ease of representation, we denote backbone as column 0 and list a few paths here as an example as shown in Fig. \ref{ensemble}: (1) (\textcolor{blue}{Blue} line in Fig. \ref{ensemble}) \textcolor{blue}{  $A0 \rightarrow A1 \rightarrow A2 \rightarrow A3 \rightarrow A4$}, (2) (\textcolor{green}{Green} line in Fig. \ref{ensemble}) \textcolor{green}{ $A0 \rightarrow A1 \rightarrow A2 \rightarrow A3 \rightarrow B3 \rightarrow C3 \rightarrow C4 \rightarrow B4 \rightarrow A4$}, (3) (\textcolor{red}{Red} line in Fig. \ref{ensemble})\textcolor{red}{ $A0 \rightarrow B0 \rightarrow B1 \rightarrow B2 \rightarrow A2 \rightarrow A3 \rightarrow A4$}, (4) (\textcolor{orange}{Orange} line in Fig. \ref{ensemble}) \textcolor{orange}{$A0 \rightarrow B0 \rightarrow C0 \rightarrow D0 \rightarrow D1 \rightarrow D2 \rightarrow C2 \rightarrow B2 \rightarrow B3 \rightarrow C3 \rightarrow C4 \rightarrow B4 \rightarrow A4$}. Each path can be viewed as a variant of FCN (except that there are pooling layers in ResNet backbone).  Therefore, ShelfNet has the potential to capture more complicated features and produce higher accuracy.

The effective number of FCN paths in ShelfNet is much larger than SegNet \cite{badrinarayanan2015segnet}, which is a single encoder-decoder pair with skip connections. The total number of paths grows exponentially with the number of encoder-decoder pairs (e.g columns 0 and 2, 3 and 4 are two pairs) and the number of spatial levels (e.g., A to D in Fig. \ref{fig:shelfnet}). Not considering the effective paths generated from residual connections in backbone, for a SegNet with 4 spatial levels (A-D), the total number of FCN paths is 4; for a ShelfNet with the same spatial levels, the total number of FCN paths is 29. The unique structure of ShelfNet significantly increases the number of effective FCN paths, thus achieving a higher accuracy.  

\subsection{Ensemble of Deep and Shallow Paths}

\begin{figure}[h]
    \centering
    \includegraphics[width = \linewidth]{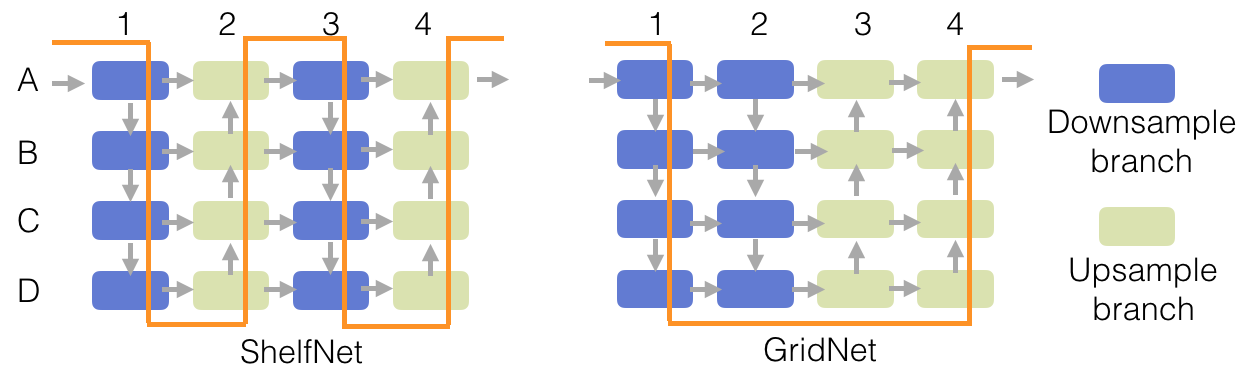}
    \caption{\small{Deepest path (orange curve) in ShelfNet and GridNet.}}
    \label{fig:grid}
\end{figure}

GridNet \cite{fourure2017residual} also has a multi-column structure. We illustrate the key difference between ShelfNet and GridNet in Fig.~\ref{fig:grid} by simplifying the structure. When stacking the same number of encoder-decoder blocks, the information path in ShelfNet can go much deeper. For example, the deepest path in ShelfNet goes through all 16 blocks, while GridNet can only use 10 blocks. The difference in depth is even bigger with more ``downsample-upsample" branch pairs. Our ShelfNet takes advantage of ensemble information from a variety of shallow and deep paths, like the success of ResNet.
\subsection{Shared-weights Residual Block (S-block)}

Compared with SegNet, the larger effective number of FCN paths comes at a price of extra blocks. To reduce the model size, we propose a modified residual block (S-Block) as shown in Fig. \ref{fig:shelfnet}(b). Here we use it only in the segmentation shelf. The two convolutional layers in the same block share the same weights while the two batch normalization layers are different. The shared-weights design that reuses weights of convolution is similar to the recurrent convolutional neural network (RCNN) \cite{alom2017inception} and recursive convolutional network \cite{kim2016deeply}.  A drop-out layer is added between two convolutional layers to avoid overfitting. The shared-weights residual block combines the strength of skip connection, recurrent convolution and drop-out regularization while  using many fewer parameters than a standard residual block.

\section{Experiments and Results}
We carried out extensive experiments to validate the fast inference speed and high accuracy of ShelfNet on three public datasets: PASCAL VOC 2012, PASCAL Context and Cityscapes. Performance is measured by mean intersection over union (mIoU). To test inference speed, we feed a single image to the network, and measure the mean running time of 100 repetitions. We first introduce the datasets and inplementation details, then provide an ablation analysis for ShelfNet, finally we compare the performance with state-of-the-art methods.


\subsection{Datasets and Implementation Details}

\subsubsection{PASCAL VOC 2012}
PASCAL VOC 2012 \cite{everingham2015pascal} contains 20 object classes with one background class. We use the augmented PASCAL VOC dataset \cite{hariharan2015hypercolumns} containing 10582,  1449 and 1456 images for training, validation and test set. MS COCO \cite{lin2014microsoft} is also used as extra training data to generate higher accuracy. 

\subsubsection{PASCAL-Context}

PASCAL-Context dataset \cite{mottaghi2014role} provides dense labels for the whole image with 59 classes and a background class. There are 4,998 training images and 5,105 test images. 

\subsubsection{Cityscapes}

Cityscapes \cite{cordts2016cityscapes} consists of images for 50 cities in different seasons and are annotated with 19 categories. It contains 2975, 500 and 1525 fine-labeled images for training, validation and test respectively. More than 20,000 images with coarse annotations are also provided.

\subsubsection{Implementation Details}
\label{sec_implementation}
All models are implemented with PyTorch \cite{paszke2017automatic}. Learning rate is scheduled in the form $lr = baselr \times (1-\frac{iter}{total\_iter})^{power}$ with $power=0.9$ as in \cite{zhang2018context}, and cross-entropy loss is used. The weight-decay is set as $10^{-4}$.  For Pascal VOC, the model is first trained with Stochastic Gradient Descent (SGD) optimizer on MS COCO dataset for 30 epochs with a base learning rate of 0.01, then trained on PASCAL augmented dataset for 50 epochs with a base learning rate of 0.001, and finally fine-tuned on original PASCAL VOC dataset for 50 epochs with a base learning rate of 0.0001. For PASCAL-Context, we train our model with SGD optimizer for 80 epochs with cross-entropy loss. The base learning rate is set as 0.001. For Cityscapes dataset, we train the model using fine-labelled images, with an initial learning rate of 0.01. 

For data augmentation, the image is randomly flipped and scaled between 0.5 to 2. The images are also randomly rotated between -10 and 10 degrees. The image is cropped into size $512 \times 512$ for PASCAL VOC and PASCAL Context datasets, and cropped into $1536\times 768$ for Cityscapes dataset. For final prediction results, we predict the segmentation masks on multi-scale inputs ranging from 0.5 to 2 and calculate their average.
\subsection{Ablation Analysis}
\subsubsection{Ablation Study for Structure}
We conduct the ablation analysis on PASCAL VOC validation dataset, and train all models with the same strategy as in Sec.~\ref{sec_implementation}. The results are summarized in Table \ref{table_structure}. Numbers after model name represent which column in Fig.~\ref{fig:shelfnet} is present. 

First we show the shelf-shaped structure, which increases the number of paths for information flow, is the key to high performance. Compared with FCN, SegNet uses columns 0,1 and 2, and uses low-level features, thus generating a higher mIoU. We use W-Net which has a similar structure compared to ShelfNet, except skip connections between (B2,C2) and (B3,C3) are \textbf{removed} in W-Net. W-Net is equivalent to stacked hourglass network (SHN) \cite{newell2016stacked}. The number of paths for information flow from input to output has the following order: ShelfNet$>$W-Net$>$ SegNet$>$FCN, and we observe the same order of segmentation accuracy, which validates our argument.

We further validate that the improvement in accuracy comes from increased number of paths, not increased number of parameters. We reduce the number of channels in a ShelfNet by half (e.g. reduce from 64 to 32 channels for A2), but keep the backbone unchanged. This reduces the parameter number from 38.7M to 29.2M, but generates a comparable result (mIoU drops from 88.26\%  to 85.52\%), and the result is much better compared to FCN with more parameters (35.0M).
\begin{table}[h]
\scalebox{0.63}{
\begin{tabular}{l|l|l|l|l|l}
\hline
Backbone & Model    & A2 \# Channels   & \# Parameter & mIoU  & Accuracy\\ \hline
ResNet50 & FCN (0)    &                   &     35.0M         &   66.11    &  91.0   \\
ResNet50 & SegNet (0,1,2)    & 64                & 31.7M          & 86.90 & 96.77   \\
ResNet50 & ShelfNet (0,1,2,3,4) & 32                & 29.2M          & 85.52 & 96.37   \\
ResNet50 & W-Net (0,1,2,3,4)& 64                &  38.7M          & 86.98 & 96.79   \\
ResNet50 & ShelfNet (0,1,2,3,4) & 64                &  38.7M         & \textbf{88.26} & \textbf{97.09}   \\
Xception & ShelfNet (0,1,2,3,4) & 64                &  35.9M          & 85.99 & 96.54   \\ \hline
\end{tabular}
}
\caption{\small{Comparison of different structures on PASCAL VOC validation dataset. Numbers in the column named ``Model" represents which column in Fig.~\ref{fig:shelfnet} is present, and column 0 represents the backbone. For example, FCN only has a single column 0 followed by a convolution layer, SegNet  has columns 0, 1 and 2, ShelfNet has columns 0 to 4 \textbf{with} skip connections between column 2 and 3, while W-Net has columns 0 to 4 \textbf{without} skip connections between (B2,C2) and (B3,C3).}}
\label{table_structure}
\end{table}
\begin{table}[]
\begin{tabular}{l|l|l|l}
\hline
Share-weights             & \# Parameter & mIoU  & Accuracy \\ \hline
                          &       45.8M       & 88.02 & 97.01    \\
\checkmark &    38.7M          & 88.26 & 97.09    \\ \hline
\end{tabular}
\caption{\small{ Comparison of ShelfNet with ResNet50 backbone using shared-weights and conventional residual blocks on PASCAL VOC 2012 validation dataset.}
}
\label{table_share_weight}
\end{table}
\subsubsection{Shared-weights Blocks (S-Block)}
\label{sec_share_weight}
We evaluate the effect of Shared-weights blocks (S-blocks) by replacing S-blocks in the segmentation shelf with not-shared-weight version of residual blocks. Both versions use the same ResNet50 backbone provided by PyTorch official website \cite{paszke2017automatic}. Both models are trained with SGD optimizer on PASCAL augmented dataset for 50 epochs with a base learning rate of 0.01, and finally fine-tuned on original PASCAL VOC 2012 dataset for 50 epochs with a base learning rate of 0.001. We test them on the PASCAL VOC validation dataset, and results are summarized in Table~\ref{table_share_weight}. The shared-weight strategy reduces the number of parameters from 45.8M to 38.7M without sacrificing the accuracy.  In our experiment the shared-weights strategy even generates a slightly higher mIoU (88.26\%) compared to the conventional residual block (88.02\%).


\subsubsection{Speed Analysis for Backbone}
ShelfNet is a flexible architecture and can be used with various backbone models. We summarize the inference speed of different backbones in Table \ref{table_backbone}, and show that dilated convolution significantly reduces the inference speed. 
``Dilated" means the network uses dilated convolution to increase receptive field instead of a pooling layer. The number of flops of a convolution layer is calculated as:
\begin{equation}
Computation = C_1 \times C_2 \times K_1 \times K_2 \times H \times W
\end{equation}
where $C_1,C_2$ are channel numbers of the input and output tensors, $K_1,K_2$ are sizes of the kernel, and $H,W$ are sizes of the feature map. A dilated convolution network will generate a larger $H,W$, therefore is computationally intensive. As shown in Table \ref{table_backbone}, with the same network structure, a dilated version has about $5\times$ larger computation burden.

State-of-the-art semantic segmentation networks such as DeepLabv3 and PSPNet use dilated residual network as the backbone, while ShelfNet uses an undilated version. Therefore, ShelfNet is much faster. The speed up mainly comes from computation reduction of the backbone.
\vspace{-0.1cm}
\begin{table}[htb!]
\scalebox{0.9}{
\begin{tabular}{l|l|l|l|l}
\hline
Backbone  & Dilated & \# Parameter & FPS & Flops \\ \hline
ResNet18  &                     &      11.7M        &     286            &   9.5G    \\
ResNet18  &       \checkmark    &        11.7M      &      59           &   48.2G    \\
ResNet50  &                     &      35.6M        &     100            &  21.4G \\
ResNet50  &     \checkmark      &   35.6M       &        28         &    99.8G   \\
ResNet101 &                     &     44.5M        &     56            &   40.8G    \\
ResNet101 &     \checkmark      &   44.5M       &       15      &   177.5G    \\
Xception  &                     &    22.9M         &     67            &   24.2G    \\ \hline
\end{tabular}
}
\caption{\small{Inference speed of different backbones. ``Dilated'' means the network uses dilated convolution.}}
\label{table_backbone}
\end{table}

\begin{table*}[t]
\scalebox{0.61}{
\begin{tabular}{l|llllllllllllllllllll|l|l|l}
\hline
Method      & aero & bike & bird & boat & bottle & bus  & car  & cat  & chair & cow  & table & dog  & horse & mbike & person & plant & sheep & sofa & train & tv   & mIoU & FPS & \# Param \\ \hline
FCN \cite{long2015fully}      & 76.8 & 34.2 & 68.9 & 49.4 & 60.3   & 75.3 & 74.7 & 77.6 & 21.4  & 62.5 & 46.8  & 71.8 & 63.9  & 76.5  & 73.9   & 45.2  & 72.4  & 37.4 & 70.9  & 55.1 & 62.2 & -   \\
DeepLabv2 \cite{chen2016deeplab}   & 84.4 & 54.5 & 81.5 & 63.6 & 65.9   & 85.1 & 79.1 & 83.4 & 30.7  & 74.1 & 59.8  & 79.0 & 76.1  & 83.2  & 80.8   & 59.7  & 82.2  & 50.4 & 73.1  & 63.7 & 71.6 & -   \\
CRF-RNN \cite{zheng2015conditional}    & 87.5 & 39.0 & 79.7 & 64.2 & 68.3   & 87.6 & 80.8 & 84.4 & 30.4  & 78.2 & 60.4  & 80.5 & 77.8  & 83.1  & 80.6   & 59.5  & 82.8  & 47.8 & 78.3  & 67.1 & 72.0 & -   \\
Deconvnet \cite{noh2015learning}   & 89.9 & 39.3 & 79.7 & 63.9 & 68.2   & 87.4 & 81.2 & 86.1 & 28.5  & 77.0 & 62.0  & 79.0 & 80.3  & 83.6  & 80.2   & 58.8  & 83.4  & 54.3 & 80.7  & 65.0 & 72.5 & -   \\
GCRF  \cite{vemulapalli2016gaussian}      & 85.2 & 43.9 & 83.3 & 65.2 & 68.3   & 89.0 & 82.7 & 85.3 & 31.1  & 79.5 & 63.3  & 80.5 & 79.3  & 85.5  & 81.0   & 60.5  & 85.5  & 52.0 & 77.3  & 65.1 & 73.2 & -   \\
DPN  \cite{liu2015semantic}       & 87.7 & 59.4 & 78.4 & 64.9 & 70.3   & 89.3 & 83.5 & 86.1 & 31.7  & 79.9 & 62.6  & 81.9 & 80.0  & 83.5  & 82.3   & 60.5  & 83.2  & 53.4 & 77.9  & 65.0 & 74.1 & -   \\
Piecewise \cite{lin2016efficient}  & 90.6 & 37.6 & 80.0 & 67.8 & 74.4   & 92.0 & 85.2 & 86.2 & 39.1  & 81.2 & 58.9  & 83.8 & 83.9  & 84.3  & 84.8   & 62.1  & 83.2  & 58.2 & 80.8  & 72.3 & 75.3 & -   \\
ResNet38 \cite{wu2016wider}   & 94.4 & 72.9 & 94.9 & 68.8 & 78.4   & 90.6 & 90.0 & 92.1 & 40.1  & 90.4 & 71.7  & 89.9 & 93.7  & 91.0  & 89.1   & 71.3  & 90.7  & 61.3 & 87.7  & 78.1 & 82.5 & 13 & 55.9M \\
PSPNet  \cite{zhao2017pyramid}    & 91.8 & 71.9 & 94.7 & 71.2 & 75.8   & 95.2 & 89.9 & 95.9 & 39.3  & 90.7 & 71.7  & 90.5 & 94.5  & 88.8  & 89.6   & 72.8  & 89.6  & 64.0 & 85.1  & 76.3 & 82.6 & 11 & 67.6M  \\
EncNet  \cite{zhang2018context}    & 94.1 & 69.2 & 96.3 & 76.7 & 86.2   & 96.3 & 90.7 & 94.2 & 38.8  & 90.7 & 73.3  & 90.0 & 92.5  & 88.8  & 87.9   & 68.7  & 92.6  & 59.0 & 86.4  & 73.4 & 82.9 & 12 & 54.5M \\ \hline
ShelfNet18 & 81.6 & 56.7 & 89.7 & 62.3 & 69.6 & 88.5 & 82.8 & 88.4 & 30.5 & 82.1 & 63.5 & 80.9 & 82.3 & 82.6 & 81.2 & 62.1 & 81.3 & 55.4 & 75.2 & 62.7 & \textbf{74.0} & \textbf{103} & 23.5M\\ 
ShelfNet50  & 94.0 & 63.2 & 86.1 & 68.9 & 73.3   & 93.6 & 87.7 & 91.5 & 31.4  & 87.1 & 67.9  & 89.5 & 88.8  & 86.2  & 85.5   & 69.9  & 88.5  & 56.1 & 82.4  & 72.3 & \textbf{79.0} & \textbf{59}  & 38.7M\\ 
ShelfNet101 & 93.6 & 64.2 & 86.9 & 69.7 & 76.2   & 93.4 & 90.5 & 94.4 & 37.0  & 91.7 & 71.1  & 91.2 & 91.5  & 88.9  & 86.2   & 72.7  & 92.6  & 58.5 & 85.8  & 72.4 & \textbf{81.1} & \textbf{42}  & 57.7M\\ \hline
\end{tabular}
}
\caption{Results on PASCAL VOC test set \textbf{without} pre-training on COCO. ShelfNet with ResNet18, ResNet50 and ResNet101 as backbone are named as ShelfNet18, ShelfNet50 and ShelfNet101 respectively. We implemented several models and measured the inference speed on a $512 \times 512$ image as input with a single GTX 1080Ti GPU. }
\label{voc_table}
\end{table*}

\begin{table*}[t]
\scalebox{0.61}{
\begin{tabular}{l|llllllllllllllllllll|l|l|l}
\hline
Method      & aero  & bike  & bird & boat & bottle & bus  & car  & cat  & chair & cow  & tale & dog  & horse & mbike & person & plant & sheep & sofa & train & tv      & mIoU & FPS & \# Param\\ \hline
CRF-FCN   \cite{zheng2015conditional}  & 90.4  & 55.3  & 88.7 & 68.4 & 69.8   & 88.3 & 82.4 & 85.1 & 32.6  & 78.5 & 64.4 & 79.6 & 81.9  & 86.4  & 81.8   & 58.6  & 82.4  & 53.5 & 77.4  & 70.1    & 74.7 & -   \\
Dilation8  \cite{yu2015multi} & 91.7  & 39.6  & 87.8 & 63.1 & 71.8   & 89.7 & 82.9 & 89.8 & 37.2  & 84   & 63   & 83.3 & 89    & 83.8  & 85.1   & 56.8  & 87.6  & 56   & 80.2  & 64.7    & 75.3 & -   \\
DPN    \cite{liu2015semantic}      & 89    & 61.6  & 87.7 & 66.8 & 74.7   & 91.2 & 84.3 & 87.6 & 36.5  & 86.3 & 66.1 & 84.4 & 87.8  & 85.6  & 85.4   & 63.6  & 87.3  & 61.3 & 79.4  & 66.4    & 77.5 & -   \\
Piecewise  \cite{lin2016efficient} & 94.1  & 40.7  & 84.1 & 67.8 & 75.9   & 93.4 & 84.3 & 88.4 & 42.5  & 86.4 & 64.7 & 85.4 & 89    & 85.8  & 86     & 67.5  & 90.2  & 63.8 & 80.9  & 73      & 78.0 & -   \\
DeepLabv2 \cite{chen2016deeplab}   & 92.6  & 60.4  & 91.6 & 63.4 & 76.3   & 95   & 88.4 & 92.6 & 32.7  & 88.5 & 67.6 & 89.6 & 92.1  & 87    & 87.4   & 63.3  & 88.3  & 60   & 86.8  & 74.5    & 79.7 & 12 & 43.9M  \\
RefineNet  \cite{lin2017refinenet} & 95    & 73.2  & 93.5 & 78.1 & 84.8   & 95.6 & 89.8 & 94.1 & 43.7  & 92   & 77.2 & 90.8 & 93.4  & 88.6  & 88.1   & 70.1  & 92.9  & 64.3 & 87.7  & 78.8    & 83.4 & 14 &118M \\
ResNet38  \cite{wu2016wider}  & 96.2  & 75.2  & 95.4 & 74.4 & 81.7   & 93.7 & 89.9 & 92.5 & 48.2  & 92   & 79.9 & 90.1 & 95.5  & 91.8  & 91.2   & 73    & 90.5  & 65.4 & 88.7  & 80.6    & 84.9 & 13  & 55.9M\\
PSPNet  \cite{zhao2017pyramid}    & 95.8  & 72.7  & 95   & 78.9 & 84.4   & 94.7 & 92   & 95.7 & 43.1  & 91   & 80.3 & 91.3 & 96.3  & 92.3  & 90.1   & 71.5  & 94.4  & 66.9 & 88.8  & 82      & 85.4 & 11 & 67.6M \\
DeepLabv3 \cite{chen2017rethinking}  & 96.4  & 76.6  & 92.7 & 77.8 & 87.6   & 96.7 & 90.2 & 95.4 & 47.5  & 93.4 & 76.3 & 91.4 & 97.2  & 91    & 92.1   & 71.3  & 90.9  & 68.9 & 90.8  & 79.3    & 85.7 & 8   & 58.0M\\
EncNet   \cite{zhang2018context}   & 95.3  & 76.9  & 94.2 & 80.2 & 85.2   & 96.5 & 90.8 & 96.3 & 47.9  & 93.9 & 80   & 92.4 & 96.6  & 90.5  & 91.5   & 70.8  & 93.6  & 66.5 & 87.7  & 80.8    & 85.9 & 12  & 54.5M \\ \hline
ShelfNet18 & 92.7 & 64.4 & 91.8 & 72.3 & 76.0 & 90.6 &84.4 & 91.1 & 34.8 & 89.5 & 68.2 & 83.6 & 88.1 & 86.8 & 85.5 & 70.6 & 85.6  & 62.0 & 83.5 & 68.7 & \textbf{79.3} & \textbf{103} & 23.5M \\
ShelfNet50  & 95.6 & 71.5 & 94.2 & 72.4 & 74.3   & 94.1 & 88.4 & 92.6 & 35.6  & 93.9 & 77.8 & 88.2 & 95.5  & 89.7  & 88.7   & 71.3  & 91.4  & 61.6 & 87.9  & 77.1    & \textbf{82.8} & \textbf{59} & 38.7M \\ 
ShelfNet101 & 95.4 & 73.9 & 94.9 & 75.7 & 83.2   & 96.3 & 91.2 & 93.9 & 35.3  & 90.0 & 79.4 & 90.2 & 94.2  & 92.8  & 90.1   & 73.2  & 92.3  & 64.5 & 88.0  & 77.5 & \textbf{84.2} & \textbf{42}  & 57.7M\\ \hline
\end{tabular}
}
\caption{Results on PASCAL VOC test set \textbf{with} pre-training on COCO.}
\label{voc_table_coco}
\end{table*}

\begin{table*}[h!]
\scalebox{0.75}{
\begin{tabular}{l|llllllll}
\hline
Model     & RefineNet-101 & RefineNet-152 & RefineNet-LW-50 & RefineNet-LW-101 & RefineNet-LW-152 & ShelfNet-18 & ShelfNet-50   & ShelfNet-101  \\ \hline
mIoU, \%  & 82.4          & 83.4          & 81.1            & 82.0             & 82.7     & \textbf{79.3}        & \textbf{82.8} & \textbf{84.2} \\
FPS       & 19            & 16            & 53              & 37               & 29          &\textbf{103}     & \textbf{59}   & \textbf{42}  \\ \hline
\end{tabular}
}
\caption{\small{Results on PASCAL VOC test set. Comparison with RefineNet and Lightweight-RefineNet (RefineNet-LW). Numbers represent the number of layers in backbone ResNet.}}
\label{voc_table_speed}
\end{table*}

\subsection{Results}
\subsubsection{PASCAL VOC 2012}
\label{pascal_voc_section}
\begin{figure}[htb!]
    \centering
        \includegraphics[width=0.9\linewidth]{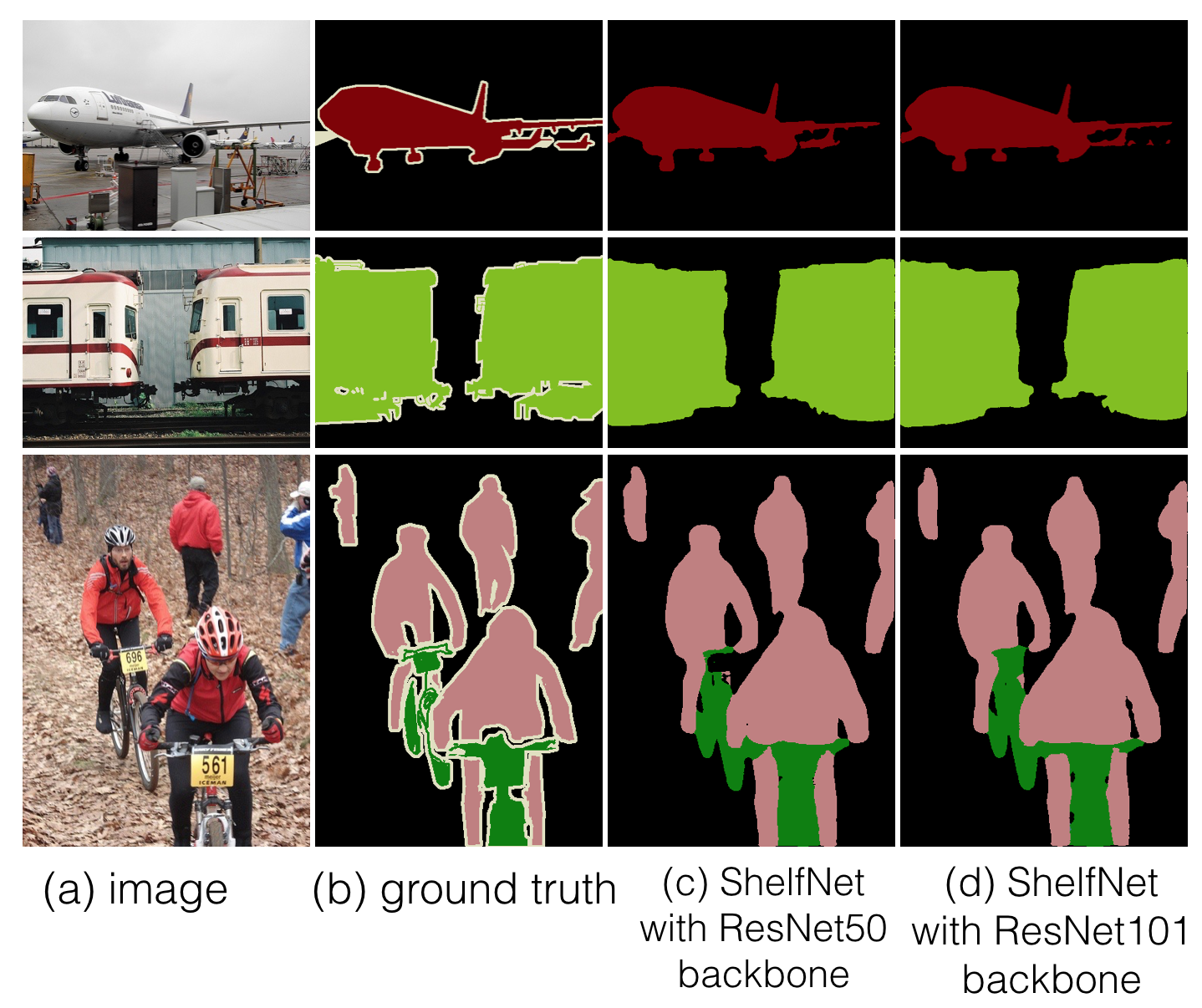}
        \caption{Examples on PASCAL VOC. Columns from left to right represent: input images, ground truth annotations, predictions from ShelfNet with ResNet50 backbone, predictions from ShelfNet with ResNet101 backbone.}
    \label{fig:pascal_result}
\end{figure}

\paragraph{Comparison with non real-time models}
We evaluate the segmentation results on the PASCAL evaluation server. Exemplary results are shown in Fig.~\ref{fig:pascal_result}. The detailed results are summarized in Table~\ref{voc_table} and Table~\ref{voc_table_coco}. For a fair comparison, we implemented ShelfNet and several state-of-the-art segmentation models with PyTorch and measured their inference speed on a single GTX 1080Ti GPU. ShelfNet with ResNet18, ResNet50 and ResNet101 backbone are named as ShelfNet18, ShelfNet50 and ShelfNet101 for short respectively. When trained only on augmented PASCAL training set and fine-tuned on original PASCAL VOC dataset, ShelfNet18, ShelfNet50 and ShelfNet101 achieve 74.0\%, 79.0\% and 81.1\% mIoU respectively. When trained on both MS COCO and PASCAL dataset,  they achieve 79.3\%, 82.8\% and 84.2\% mIoU respectively. We provide anonymous links to our results\footnote{\url{http://host.robots.ox.ac.uk:8080/anonymous/5NMB0K.html}}\footnote{\url{http://host.robots.ox.ac.uk:8080/anonymous/KAZMJD.html}}.

Compared to state-of-the-art semantic segmentation models such as PSPNet \cite{zhao2017pyramid} and EncNet \cite{zhang2018context}, ShelfNet achieves a comparable mIoU but generates 4 to 5 times speed-up during inference (59 FPS for ShelfNet50 and 42 FPS for ShelfNet101, 11 FPS for PSPNet and 12 FPS for EncNet). Compared to DeepLabv2, our ShelfNet achieves comparable mIoU (79.3\% vs 79.7\%) but achieve almost $10\times$ speed up (103FPS vs 12 FPS).  Compared with large networks such as PSPNet and DeepLab, our ShelfNet is faster mainly because there's no dilation in the backbone, thus less computation, as shown in Table~\ref{table_backbone}.

\paragraph{Comparison with real-time models}
Lightweight-RefineNet (LWRF) \cite{nekrasov2018light} is one of the state-of-the-art real-time semantic segmentation models. Comparisons between our ShelfNet and LWRF are summarized in Table \ref{voc_table_speed}. ShelfNet50 achieves higher accuracy (82.8\%) than LWRF with a ResNet 152 backbone (82.7\%) and RefineNet with a ResNet101 backbone (82.4\%). Compared to RefineNet and LWRF, the better performance with a much smaller backbone of ShelfNet validates the efficiency of the proposed shelf-like structure in feature extraction. Our ShelfNet with Resnet101 backbone achieves the highest accuracy (84.2\%) compared to all RefineNet and LWRF models. 

In addition to the higher accuracy, ShelfNet achieves faster inference speed compared with LWRF when using the same backbone. Unlike LWRF, our ShelfNet reduces channel number with a $1 \times 1$ convolution as in column 1 in Fig.~\ref{fig:shelfnet}. For levels A to D, LWRF uses 256, 256, 256, 512 channels, while in ShelfNet it's reduced to 64, 128, 256, 512. As mentioned in Sec.~\ref{sec:channel_reduction}, the computation for a conv layer is $H \times W \times K^2 \times C^2$,hence quadratically reduced to $(\frac{64}{256})^2=\frac{1}{16}$ for level A; even with two more columns, overall the computation burden is reduced to $2 \times \frac{1}{16}=\frac{1}{8}$. Therefore, our ShelfNet is faster.

\begin{figure}[htb!]
    \centering
        \includegraphics[width=0.9\linewidth]{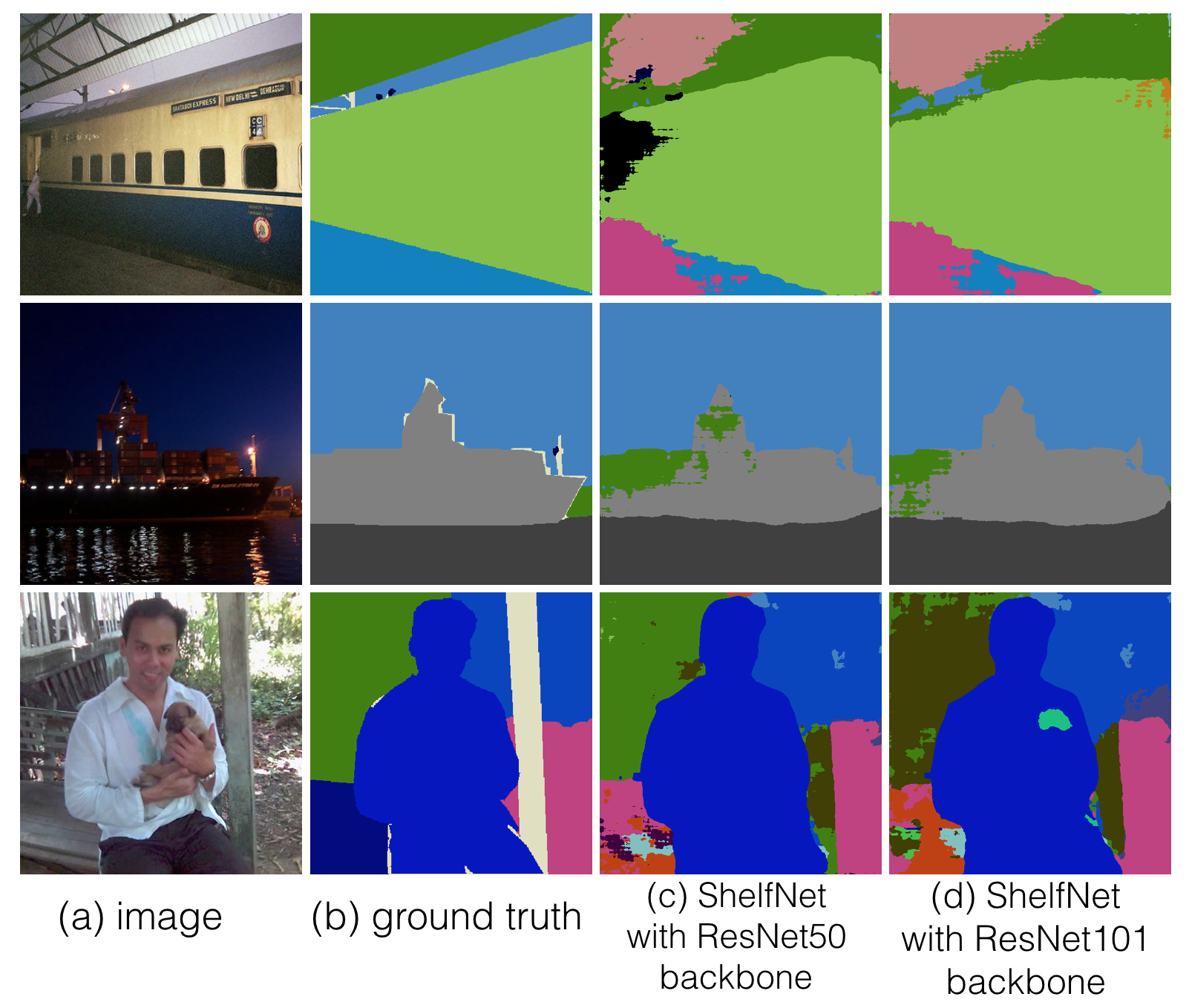}
        \caption{Example predictions of ShelfNet on PASCAL Context  dataset.}
    \label{context_result}
    \label{context_figure}
\end{figure} 
\vspace{-0.1cm}

\subsubsection{PASCAL-Context}
\label{pcontext_section}

Examples of ShelfNet on PASCAL-Context test set are shown in Fig.~\ref{context_figure}. The detailed results are summarized in Table \ref{context_table}. DeepLab-v2 achieves 45.7\% mIoU with MS COCO as extra training data, while our ShelfNet achieves 45.6\% and 48.4\% with ResNet50 and ResNet101 respectively without extra training data. RefineNet achieves 47.3\% mIoU at the speed of 29 FPS, while our ShelfNet achieves 45.6\% mIoU at 59 FPS with ResNet50 backbone, and 48.4\% mIoU at 42 FPS with ResNet100 backbone. ShelfNet has both higher accuracy and faster running speed compared with RefineNet. EncNet achieves a higher mIoU of 51.7\%; this is because EncNet uses dilated convolution and sacrifices the inference speed. The inference speed of ShelfNet is 4 to 5 times faster than EncNet as shown in Table \ref{voc_table_coco}. Overall, our ShelfNet achieves high mIoU with fast inference speed.
\begin{figure*}[htb]
    \centering
        \includegraphics[width=0.6\linewidth]{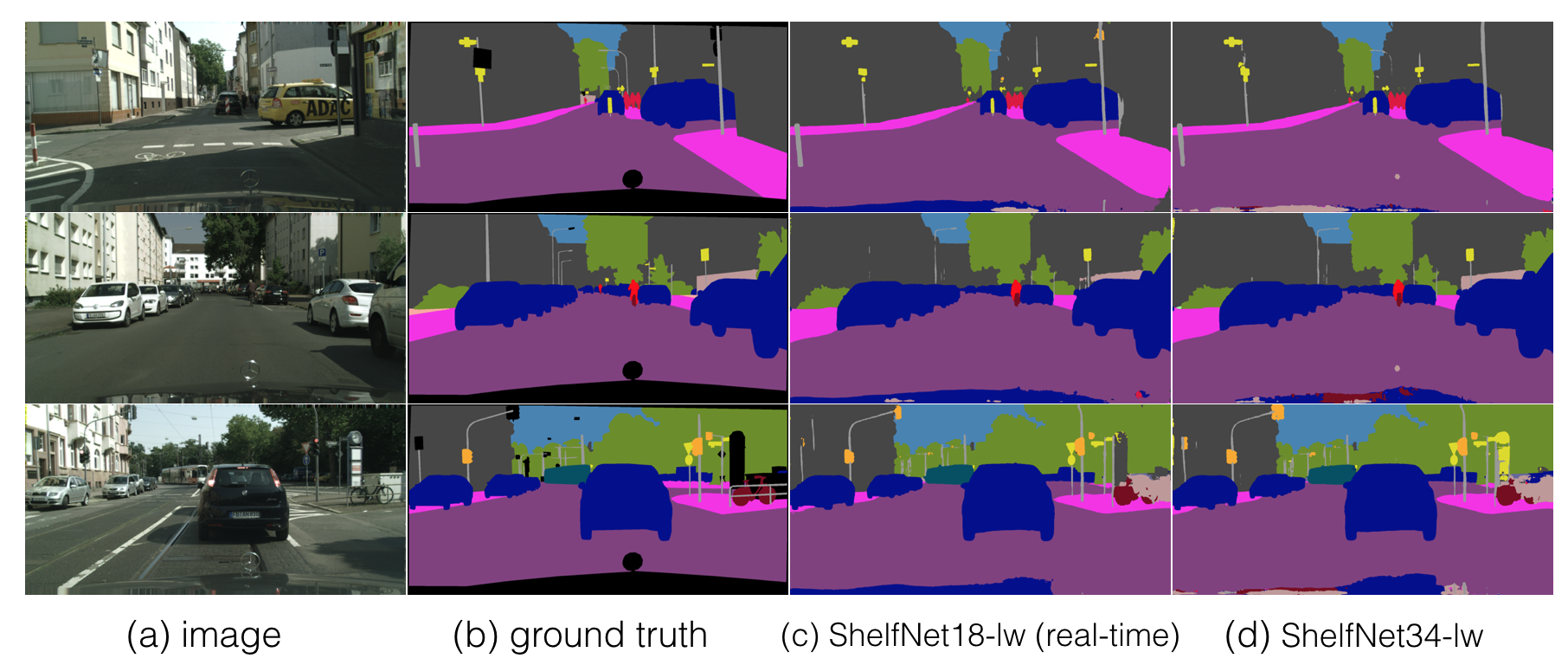}
        \caption{\small{Results of ShelfNet on Cityscapes validation dataset. Results from real-time models are generated from single scale test.}
        }
    \label{cityscape_result}
    \label{city_figure}
\end{figure*} 
\vspace{-0.4cm}
\begin{table}[htb!]
\centering
\scalebox{0.8}{
\begin{tabular}{l|l|l|l|l}
\hline
Model      & BaseNet     & mIoU, \% & FPS & \# Parameters\\ \hline
FCN-8s \cite{long2015fully}    &             & 37.8     &  &\\
CRF-RNN \cite{zheng2015conditional}   &             & 39.3   &  &\\
ParseNet \cite{liu2015parsenet}  &             & 40.4     &  & \\
Piecewise \cite{lin2016efficient} &             & 43.3     & & \\
DeepLab-v2\cite{chen2016deeplab} & Res101-COCO & 45.7     & 12 & 43.9M \\
RefineNet \cite{lin2017refinenet} & Res101      & 47.1 & 19 & 118M\\
RefineNet \cite{lin2017refinenet} & Res152      & 47.3 &  16 & 134M  \\ 
EncNet  \cite{zhang2018context}    &          & 51.7 & 12 &54.5M\\  \hline
ShelfNet50   & Res50       & \textbf{45.6} & 59  & 38.7M \\
ShelfNet101   & Res101      & \textbf{48.4} & 42 & 57.7M \\ \hline
\end{tabular}
}
\caption{Segmentation results on PASCAL-Context  dataset}
\label{context_table}
\end{table}
\vspace{-0.8cm}
\subsubsection{Cityscapes}

For non real-time tasks, we used the same structure as in Fig.~\ref{fig:shelfnet}, which is used for experiments on PASCAL datasets. For real-time tasks, we modified the network for a light-weight structure, denoted as ShelfNet-lw.

\paragraph{Light-weight ShelfNet}
We further modify our ShelfNet into a light-weight version (ShelfNet-lw). Different from Fig.~\ref{fig:shelfnet}, in ShelfNet-lw, only features from levels B, C, D are fed into the segmentation shelf; features in level A are not used. For ShelfNet34-lw, we set the channel number for B, C, D as 128, 256, 512 respectively; for ShelfNet18-lw, we set the channel number for B, C, D as 64, 128, 256 respectively, further reducing the computation burden. We replace the conv transpose layer with a direct upsampling to reduce number of parameters, and replace S-block in the upsample branch with a conv-bn-relu along with channel attention. For training of our light-weight ShelfNet, we use online hard example mining loss and distributed training to match the training scheme for BiSeNet. 
\vspace{-0.1cm}
\begin{table}[h]
    \begin{subtable}{0.45\linewidth}
        \scalebox{0.7}{
\begin{tabular}{l|l}
\hline
Method & mIoU (test)                     \\ \hline
EffiConv\cite{romera2017efficient} & 68.0 \\
ICNet\cite{zhao2017icnet}                              & 69.5                             \\
ENet\cite{paszke2016enet}                             & 58.3                             \\
GUN\cite{mazzini2018guided}                                 & 70.4                            \\
ContextNet\cite{poudel2018contextnet}                         & 66.1                            \\
BiSeNet(Res18) \cite{yu2018bisenet}               & 74.7                             \\ \hline
ShelfNet18-lw             & \textbf{74.8}                            \\ \hline
\end{tabular}
}
\caption{\small{Real-time performance}}
    \end{subtable}
    \begin{subtable}{0.5\linewidth}
        \scalebox{0.7}{
\begin{tabular}{l|lll}
\hline
Model     & Backbone      & val  & test \\ \hline
RefineNet \cite{lin2017refinenet} & Res101 & -    & 73.6 \\
DeepLabv2 \cite{chen2016deeplab} & Res101 & 71.4 & 70.4 \\
DUC \cite{wang2018understanding}       & Res152 & 76.7 & 76.1 \\
PSP \cite{zhao2017pyramid}       & Res101 & -    & 78.4 \\
BiSeNet \cite{yu2018bisenet}   & Res101 & 80.3 & 78.9 \\ \hline
ShelfNet50 & Res50 & - & 74.1 \\
ShelfNet101 & Res101 & - & 77.5 \\
ShelfNet34-lw  & Res34  & \textbf{80.0} & \textbf{79.0} \\ 
\hline
\end{tabular}
}
\caption{\small{Non real-time performance}}
    \end{subtable}
    \vspace{-0.2cm}
    \caption{\small{Mean IoU (\%) for Cityscapes dataset. Real-time ShelfNet was evaluated on single scale input. Light-weight ShelfNet is marked with -lw.}}
\label{city_accuracy}
\end{table}
\vspace{-0.5cm}

\begin{table}[h]
\scalebox{0.69}{
\begin{tabular}{l|llllll|l}
\hline
Input size                 & EffConv & ICNet & ENet & GUN  & Context & BiSeNet & ShelfNet18-lw \\ \hline
1024x2048 & -             & 30.3  & -    & 33.3 & 18.3       & 37.0           & 36.9            \\
1920x1280    & 11.4          & -     & 21.6    & -    & -          & 31.3           & 31.2            \\
768x1536  & -             & -     & -    & -    & -          & 61.7           & 59.2            \\ \hline
\end{tabular}
}
\caption{\small{Speed analysis (FPS) for a single forward pass. BiSeNet and ShelfNet use ResNet18 as backbone and are tested on a single GTX 1080Ti GPU. Speed for other models are from the literature.}}
\label{city_speed}
\end{table}
\vspace{-0.8cm}

\paragraph{Numerical results}
For non real-time tasks, we average results from multi-scale evaluations; for real-time tasks, we use single-scale evaluation.

The results are summarized in Fig.~\ref{cityscape_result}, Table~\ref{city_accuracy} and Table~\ref{city_speed}. ShelfNet50 achieves 74.1\% mIoU, and ShelfNet101 achieves 77.5\% mIoU.

Our ShelfNet achieves significant improvement in both inference speed and accuracy. Our ShelfNet18-lw achieved 74.8\% mIoU with single-scale evaluation, surpassing all existing real-time models. In terms of inference speed, our ShelfNet18-lw achieves comparable inference speed as BiSeNet, surpassing previous real-time models such as ICNet and GUN.

Our ShelfNet34-lw achieves the highest mIoU of 79.0\% on Cityscapes test set. Our model outperforms previous non real-time models with a large backbone such as ResNet101, which provides strong evidence for the effectiveness of our shelf-shaped structure. We provide links to results\footnote{\url{https://tinyurl.com/y65vt9ct}}\footnote{\url{https://tinyurl.com/y6ed2uf9}}.

\section{Conclusion}
We proposed ShelfNet for fast semantic segmentation, which has multiple pairs of encoder-decoder branches with skip connections between adjacent branches. The unique shelf-shaped structure enables multiple paths for information flow and achieves high accuracy; the shared-weight design in the S-block significantly reduces parameter number without sacrificing accuracy. We validated the high segmentation accuracy and fast running speed on three benchmark datasets. ShelfNet achieves comparable segmentation accuracy to state-of-the-art off-line models with a 4 to 5 times faster inference speed. Our real-time model achieves the highest mIoU on the Cityscapes dataset, while maintaining a high inference speed comparable to BiSeNet. Our off-line ShelfNet with ResNet34 backbone outperforms previous models with large backbones such as ResNet101, validating the effectiveness of our shelf-shaped structure.

{\small
\bibliographystyle{ieee}
\bibliography{egbib}
}

\end{document}